\documentclass[twoside]{article}

\usepackage{PRIMEarxiv}

\usepackage[utf8]{inputenc} % allow utf-8 input
\usepackage[T1]{fontenc}    % use 8-bit T1 fonts
\usepackage{hyperref}       % hyperlinks
\usepackage{url}            % simple URL typesetting
\usepackage{booktabs}       % professional-quality tables
\usepackage{amsfonts}       % blackboard math symbols
\usepackage{nicefrac}       % compact symbols for 1/2, etc.
\usepackage{microtype}      % microtypography
\usepackage{lipsum}
\usepackage{fancyhdr}       % header
\usepackage{graphicx}       % graphics
\graphicspath{{media/}}     % organize your images and other figures under media/ folder
\usepackage{multirow}
\usepackage{bigstrut}

\title{Connector 0.5: A unified framework for graph representation learning}

\author{
  Thanh Sang Nguyen${}^{*}$, Jooho Lee${}^{*}$, Van Thuy Hoang${}^{*}$, O-Joun Lee${}^{\dagger}$ \\
  Department of Artificial Intelligence \\
  The Catholic University of Korea \\
  Bucheon-si, Gyeonggi-do 14662, Republic of Korea\\
  \texttt{\{sangnguyen,jooho0223,hoangvanthuy90,ojlee\}@catholic.ac.kr} \\
  }

\begin{document}
\maketitle

\begin{abstract}
%\lipsum[1]

Graph representation learning models aim to represent the graph structure and its features into low-dimensional vectors in a latent space, which can benefit various downstream tasks, such as node classification and link prediction.
Due to its powerful graph data modelling capabilities, various graph embedding models and libraries have been proposed to learn embeddings and help researchers ease conducting experiments.
In this paper, we introduce a novel graph representation framework covering various graph embedding models, ranging from shallow to state-of-the-art models, namely Connector.
First, we consider graph generation by constructing various types of graphs with different structural relations, including homogeneous, signed, heterogeneous, and knowledge graphs.
Second, we introduce various graph representation learning models, ranging from shallow to deep graph embedding models.
Finally, we plan to build an efficient open-source framework that can provide deep graph embedding models to represent structural relations in graphs.
The framework is available at \url{https://github.com/NSLab-CUK/Connector}.
\let\thefootnote\relax
\footnotetext{${}^{*}$ These authors have equally contributed to this study.} %%%%%%%%%%
\footnotetext{${}^{\dagger}$ Correspondence: \texttt{ojlee@catholic.ac.kr}; Tel.: +82-2-2164-5516} %%%%%%%%%%
%\footnotetext{${}^{1}$ Both authors contributed equally to this research.} %%%%%%%%%%
\end{abstract}

% keywords can be removed

\section{Introduction}
\label{sec:intro}

Graphs are a universal language representing and visualizing relationships and connections between different entities or data points~\cite{perozzi2014deepwalk,grover2016node2vec,ou2016asymmetric}.
Graph structure widely exists in various practical application systems. 
For example, relationships between online users in social media could form a large social graph network.
Another example is the recommendation system, where users' behaviours such as purchasing, browsing and rating products can be abstracted into an interaction graph between users and products. 
Graph representation learning methods aim to learn nodes and edges of graphs as low-dimensional vectors, mainly in Euclidean space~\cite{s23084168}.
These representations could then be used directly to improve various downstream tasks, such as node classification, link prediction, and visualization tasks.

However, large amounts of data are still represented by different types of graphs, including homogeneous, heterogeneous, knowledge, and signed graphs~\cite{chen2020graph}.
Over the years, various graph embedding models have been proposed to transform graph entities into low-dimensional vectors~\cite{perozzi2014deepwalk,grover2016node2vec,ou2016asymmetric,DBLP:journals/corr/KipfW16}.
There have been various graph embedding models have been proposed, such as DeepWalk~\cite{perozzi2014deepwalk}, Node2Vec~\cite{grover2016node2vec}, HOPE~\cite{ou2016asymmetric}, GCN~\cite{DBLP:journals/corr/KipfW16}, GraphSage, and GAT.
Several libraries, such as PyTorch Geometric~\cite{DBLP:journals/corr/abs-1903-02428} and CogDL~\cite{DBLP:journals/corr/abs-2103-00959}, have been developed to facilitate deep learning on graph representation learning.
However, most existing libraries only focus on providing basic components of graph neural networks and mainly consider elementary tasks, such as node and graph classification.
These libraries do not provide shallow models and handle various types of graphs, which could be essential in specific tasks~\cite{s23084168}.

To bridge this gap, this paper provides a graph representation learning framework, namely Connector, which can provide various existing graph embedding methods. 
Connector consists of the following three processes:
\begin{itemize}
    \item This framework  can process a wide range of different types of graphs, ranging from homogeneous graphs to knowledge graphs.
    \item Various graph embedding models are introduced in this framework, including shallow models and deep graph embedding models.
    \item We provide a graph classification task to evaluate the node embeddings.
\end{itemize}

The following sections of this paper are outlined as follows. 
Section \ref{sec:2} describes our framework and its components.
Section \ref{sec:Comparision} discusses several graph embedding libraries compared to Connector. 
The last section, Section \ref{sec:conclusion}, is the conclusion and future work.

%A new taxonomy of graph embeddings is proposed, which classifies both static graph and dynamic graph methods; (2) Existing model Systematic analyses provide perspectives for understanding existing methods; (3) Four potential research directions for graph embedding are proposed.

%However, most real-world graph structures are complex and different in size. 
%A complex graph network may contain billions of nodes and edges, and different edges may represent different relationships between nodes. 
%Therefore, how to effectively model graph structure information is the focus of continuous attention in academia and industry.

%Conclusion:

%The article proposes an unsupervised multi-level learning framework GraphZoom, which can improve the quality and efficiency of existing unsupervised graph embedding methods. GraphZoom consists of the following four processes:

\section{Framework description}
\label{sec:2}

\begin{figure}[thpb]
\centering
 \includegraphics[width=0.7\linewidth]{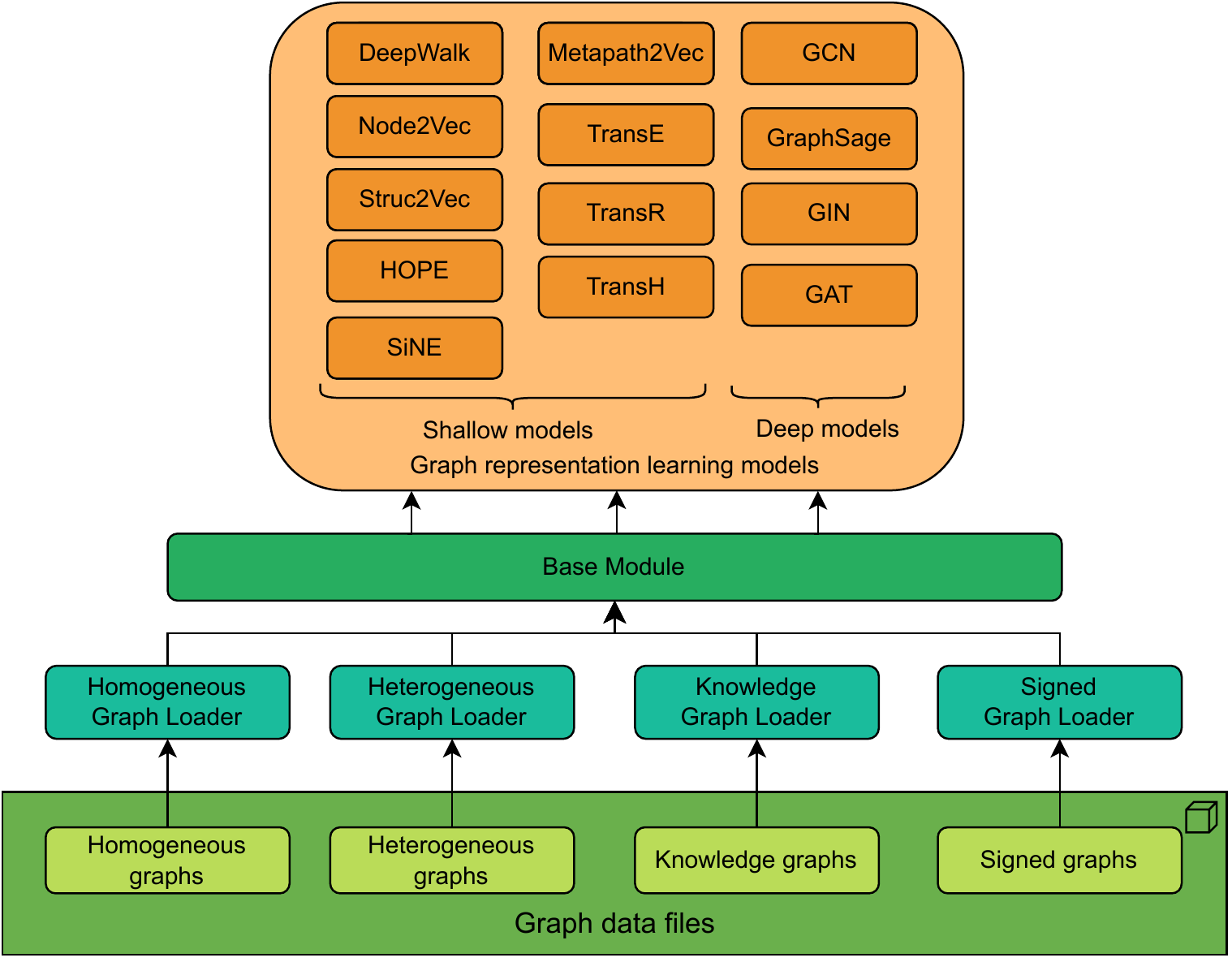}
  %\centering
  \caption{\normalfont A graphical architecture of Connector framework.
  There are three main components in Connector, including graph loader modules, base modules, and graph representation learning modules.
  }
  \label{fig:architecture}
\end{figure}

Connector is developed mainly based on Python language and PyTorch library.
Figure ~\ref{fig:architecture} presents the architecture overview of the Connector framework.
Our framework comprises three main modules: graph loaders, base model modules, and graph representation learning modules.
Graph Loader modules make it straightforward to handle different types of graphs.
Since different types of graphs are stored in different file structures, we introduce four graph loaders to handle types of graphs, including homogeneous, heterogeneous, knowledge and signed graph loaders.
There are two advantages of graph generation from stored files.
First, loading the graph data from files with Graph Loaders makes building and handling the original datasets manageable.
Second, our framework could help researchers who generated their own datasets integrate the datasets easily into our framework.
The objective of the base module is to build general tasks such as load models and load/save parameters.

\vspace{-3pt}

\begin{table}[thpb]
  \caption{A summary of benchmark datasets}
  \centering
  \label{tab:dataset}
  \begin{tabular}{l l l c c}
    \toprule
    \textbf{ \#} &\textbf{ Dataset} & \textbf{ Type }& \textbf{ $\#$ Nodes/Entities  }& \textbf{ $\#$ Edges/Relations} \\\hline
    
    1 & Karate & Homogeneous graph & 34 & 156 \\\hline
    2 & Cora & Homogeneous graph & 2,808 & 5,429\\\hline
    3 & Wiki & Homogeneous graph & 2,405 & 17,981\\\hline
    4 & Blog Catalog & Homogeneous graph & 10,312 & 333,983\\\hline
    
    5 & Eponions& Signed graph &75,879 &508,837 \\\hline
    6 & Slashdot& Signed graph &82,140 &82,140\\\hline
    
    7 & Net-dbis & Heterogeneous graph & - & -\\\hline
    8 & Net aminer & Heterogeneous graph & -  & - \\\hline
    
    9 & FB13 & Knowledge graph & 13 &75,043\\\hline
    10 & FB15k & Knowledge graph & 14,951 &592,213 \\\hline
    11 & FB15K237 & Knowledge graph &14,541 &310,116	\\\hline
    12 & NELL-995& Knowledge graph & 63,361&200 \\\hline
    13 & WN11 & Knowledge graph & 38,588 & 138,887 \\\hline
    14 & WN18 & Knowledge graph & 40,943 &151,442	 \\\hline
    15 & WN18RR& Knowledge graph &86,835	 &93,003\\\hline
    16 & YAGO3-10 & Knowledge graph & 123,182 & 37\\\hline
    
        \end{tabular}
\end{table}

\begin{itemize}
    \item \textbf{ Graph loader modules}:
    Since various types of graphs are stored in different styles, we first built a graph loader module to handle different graph file structures into NetworkX types.
    To this extent, our framework could easily incorporate into other libraries, such as PyG~\cite{DBLP:journals/corr/abs-1903-02428} and CogDL~\cite{DBLP:journals/corr/abs-2103-00959}
    There are three main data loaders: the homogeneous graph module handling edge list files in homogeneous graphs, the heterogeneous graph model handling heterogeneous graphs, and the knowledge module handling knowledge graph files.
    For heterogeneous graphs, we first extract the lists of entities and their relation and then create a data frame.
    After that, we can utilise this data frame in the NetworkX package to represent the  knowledge graphs.
    Table \ref{tab:dataset} summarises various benchmark datasets implemented in our framework.
    
    \item \textbf{ Graph representation learning models}:
    In Connector, there are various graph embedding models, including shallow models and deep models.
    Shallow models aim to learn node embeddings in the homogeneous graph by representing each node as one-hot encoding and learning an embedding matrix of all nodes.
    For deep models, we employ several GNN variants, such as GCN, GraphSage, and GAT.
    Currently, we deliver a set of graph embedding models in Connector in Table \ref{tab:model}.
      
\end{itemize}

\begin{table}
  \caption{A summary of graph embedding models introduced in Connector.}
  \centering
  \label{tab:model}
  \begin{tabular}{p{0.7 cm}p{2.8 cm}p{3 cm}p{3 cm}p{3 cm}}
    \toprule
    \textbf{ \#} &\textbf{ Model} & \textbf{ Model type }& \textbf{Graph type  }\\\hline
    
   1 &  Node2Vec~\cite{perozzi2014deepwalk} & Shallow model & Homogeneous graph \\\hline
    2 & DeepWalk \cite{grover2016node2vec} & Shallow model & Homogeneous graph \\\hline
    3 & Struct2Vec \cite{ribeiro2017struc2vec}& Shallow model & Homogeneous graph \\\hline
    4 & HOPE \cite{ou2016asymmetric}& Shallow model & Homogeneous graph \\\hline
    
   5 &  SINE \cite{DBLP:conf/sdm/WangTACL17} & Shallow model & Signed graph \\\hline
    
    6 & Metapath2Vec \cite{dong2017metapath2vec} & Shallow model & Heterogeneous graph \\\hline
    
    7 & TransE \cite{DBLP:conf/nips/BordesUGWY13} & Shallow model & Knowledge graph \\\hline
    8 & TransR \cite{DBLP:conf/aaai/LinLSLZ15} & Shallow model & Knowledge graph \\\hline
    9 & TransH \cite{DBLP:conf/aaai/WangZFC14} & Shallow model & Knowledge graph \\\hline
    
    10 & GCN \cite{DBLP:journals/corr/KipfW16} & Deep model & Homogeneous graph \\\hline
    11 & GraphSAGE \cite{DBLP:conf/nips/HamiltonYL17} & Deep model & Homogeneous graph \\\hline
    12 & GIN \cite{DBLP:conf/iclr/XuHLJ19} & Deep model & Homogeneous graph \\\hline
    13 & GAT \cite{https://doi.org/10.48550/arxiv.1710.10903} & Deep model & Homogeneous graph \\\hline
        \end{tabular}
\end{table}

%%\begin{table}[htbp]
%\caption{A Nice Table}
%\label{Tab:SRNRValues}
%\begin{center}
%\begin{tabular}{p{2.5 cm}p{3 cm}p{3 cm}}
%\hline
%\multirow{2}{*}{Model}& \multicolumn{2}{p{6cm}}{\centering Model type} \bigstrut \\
%\cline{2-3} & \multicolumn{1}{c}{Shallow model} & \multicolumn{1}{c}{Deep model} \bigstrut \\ \hline
%DeepWalk & 27.54 & 13.23 \bigstrut \\

%\hline
%%\end{tabular}
%\end{center}
%\end{table}

%\begin{tabular}{l ccccc ccccc}
%\centering
%\toprule
% & \multicolumn{5}{c}{Females} & \multicolumn{5}{c}{Males} \\
%\cmidrule(lr){2-6} \cmidrule(lr){7-11}
%Treatment     & V1   & V2  & V3 & V4 & V5 & V1   & V2  & V3 & V4 & V5 \\
%%\midrule
%Placebo       & 0.21 & 163 & 3  & 4  & 5  & 0.22 & 164 & 3  & 4  & 5  \\
%ACE Inhibitor & 0.13 & 142 & 3  & 4  & 5  & 0.15 & 144 & 3  & 4  & 5  \\
%Hydralazine   & 0.17 & 143 & 3  & 4  & 5  & 0.16 & 140 & 3  & 4  & 5  \\
%\bottomrule
%\end{tabular}

\section{Comparison with other frameworks}
\label{sec:Comparision}

Over the years, several frameworks have provided various graph representation learning models with various sampling strategies and downstream tasks. 
PyTorch Geometric (PyG)~\cite{DBLP:journals/corr/abs-1903-02428} is a library for deep learning on graphs built on top of PyTorch, designed to make it easy to work with graph-structured data using PyTorch.
PyG provides a set of efficient data loaders and utilities to preprocess, transform and augment graph data. This can save time and effort when working with large-scale graph data.
Deep Graph Library (DGL)~\cite{wang2019dgl} is a library for building and training graph neural networks. DGL provides a flexible framework for building graph neural networks, allowing users to create custom models that can be tailored to specific tasks or datasets. 
DGL is designed to work efficiently with large-scale graphs, leveraging CPU and GPU parallelism to accelerate training and inference. 
OpenNE (Open Network Embedding) is an open-source Python library for network embedding, which is a technique for learning low-dimensional vector representations of nodes in a network. 
However, the OpenNE library only introduces shallow models applied to homogeneous graphs.
Compared to OpenNE, Connector can handle various types of graphs and introduce deep models.

\section{Conclusion and future work}
\label{sec:conclusion}
In this work, we have delivered Connector, a graph representation framework covering various graph embedding models.
This framework is designed to handle various types of graphs and contains shallow models as well as deep models.
In future work, we plan to mainly focus on introducing deep graph embedding models which understand the graph structure, such as local/global positional encoding and structural encodings.
Positional encoding is a strategy that captures nodes' relative and absolute graph positions. 
In graphs, each node is commonly associated with a set of features, which can be used to represent the node in the graph. However, the position of a node in the graph is not represented in these features.
There are several strategies to represent positional encoding to graph embedding models, including adding to node features as biases or combining with message information from neighbours nodes to target nodes.
Therefore, we plan to integrate general, powerful GNNs and graph transformer models which could capture graph structure into Connector.

%In future work, we aim to build additional graph embedding models to handle several types of graphs, such as control flow or molecular graphs.
%Furthermore, we plan to integrate state-of-the-art graph transformer models into Connector.

\section*{Acknowledgments}
This work was supported 
in part by the National Research Foundation of Korea (NRF) grant funded by the Korea government (MSIT) (No. 2022R1F1A1065516) (O.-J.L.) 
and
in part by the Research Fund, 2022 of The Catholic University of Korea (M-2022-B0008-00153) (O.-J.L.). 

%Bibliography
\bibliographystyle{unsrt}  
\bibliography{references}

\end{document}